\documentclass[review]{elsarticle}

\usepackage{lineno,hyperref}
\usepackage{amssymb,amsmath,amsfonts}
\usepackage{graphicx,graphics,color,setspace}
\usepackage{latexsym,times,geometry,array}
\usepackage{psfig,epsfig,float,floatflt}
\usepackage{caption,subcaption,fancyhdr}
\modulolinenumbers[5]

\begin{document}

\begin{frontmatter}

\title{Honey Bee Dance Modeling in Real-time \\ using Machine Learning}


\author[mymainaddress]{Abolfazl Saghafi\corref{mycorrespondingauthor}}
\cortext[mycorrespondingauthor]{Corresponding author}
\ead{a.saghafi@usciences.edu}

\author[mysecondaryaddress]{Chris P. Tsokos}

\address[mymainaddress]{Department of Mathematics, Physics and Statistics, University of the Sciences, Philadelphia, PA}
\address[mysecondaryaddress]{Department of Mathematics and Statistics, University of South Florida, Tampa, FL}

\begin{abstract}
The waggle dance that honeybees perform is an astonishing way of communicating the location of food source. After over 60 years of its discovery, researchers still use manual labeling by watching hours of dance videos to detect different transitions between dance components thus extracting information regarding the distance and direction to the food source. We propose an automated process to monitor and segment different components of honeybee waggle dance. The process is highly accurate, runs in real-time, and can use shared information between multiple dances.
\end{abstract}


\end{frontmatter}

\linenumbers

\section{Introduction}

\begin{figure}[!ht]
\centering
\includegraphics[width=1\textwidth,height=1\textheight,keepaspectratio]{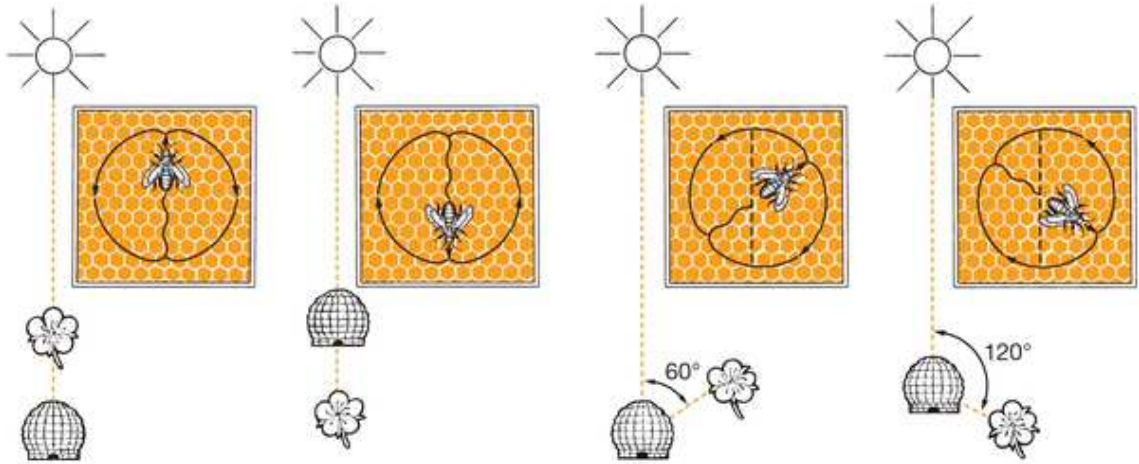}%
\caption{honeybee dance to indicate location of a food source \cite{Britannica}} \label{ch2fig1}
\end{figure}

Honey bees perform a special dance known as waggle dance within the beehive to communicate the information regarding the distance and direction of food sources. Usually, each dance pattern consists of two phases. The waggle phase during which the bee walks roughly in a straight line while rapidly shaking its body from left to right and the turning phase at the endpoint of a waggle dance in which the bee typically returns to the starting location of the waggle dance by turning in a clockwise (right) or counterclockwise (left) direction. The direction and duration of waggle dance conveys the direction and distance of the food source. For example, flowers that are located directly in line with the sun are represented by waggle runs in an upward direction on the vertical combs, and any angle to the right or left of the sun is coded by a corresponding angle to the right or left of the upward direction; see Fig. \ref{ch2fig1}. \pagebreak

The Australian scientist Karl von Frisch was the first who discovered how honeybees communicate the location of the food source through dancing \cite{Frisch} and was honored with a Nobel Prize in 1973 for his discovery. Honey bees even adjust their flight path to compensate for being blown off course by the wind and for changing position of the sun through time \cite{Riley}. However, their course is seldom so precise that they can find the food without the aid of vision and/or smell as they approach it \cite{Preece}.

Researchers have tried to model the honey bee dance patterns using various methods in order to understand the communication aspect and to mimic that pattern in building intelligent robots \cite{Feldman}. Yet after 60 years of discovering the waggle dance it is unknown how a follower bee decodes the information contained in the dance. Feldman and Balch \cite{Feldman} have used k-Nearest-Neighbors (kNN) with Hidden Markov Model (HMM) to group bees into dancer, follower, active, inactive ones based on their behaviour in the hive. Their model could achieve 78.7\% accuracy in grouping the bees while their accuracy for modeling the bee dance was 71.4\%. Xuan and Murphy \cite{Xuan} have used a first-order Auto-Regressive AR(1) model to segment honey bee dance signal into patterns of dance moves. Oh et al. \cite{Oh} have used Switching Linear Dynamic System (SLDS) in a supervised hold-one-out formulation and Parameterized Segmental SLDS which requires additional supervision during the learning process to achieve high model accuracy in segmenting dance patterns. Fox et al. \cite{Fox} have used sticky Hidden Markov Model with Hierarchical Dirichlet processes priors (HDP) and switching Vector Autoregressive (VAR) in both unsupervised and partially supervised setting to model the dance patterns. \pagebreak

Studying honey bee dance patterns requires expert-labeling of videotapes which is a time-consuming and error prone process. The objective of this study is to develop a fast and reliable system that can learn to segment and label dance patterns automatically.

\section{Materials and Methods}
\subsection{Dataset}
A dataset for honey bee dance can be found in Oh et al. \cite{Oh} where there are six set of measurements ${\mathbf x}_t=(x_t,\, y_t,\, \theta_t,\, c_t)$ in which $(x_t, y_t)$ denotes the 2D coordinates of the bee's body in $x$ and $y$ axis at time $t$, $\theta_t$ represents the bee's head angle, and $c_t$ shows the label of the dance move; $c=-1$ for turn-right, $c=0$ for turn-left, $c=1$ for waggle. The dance label has been added later investigating a video feed using TeamView software. The aim of this study is to use a machine learning approach to classify the dance patterns with high accuracy in real-time. Oh et al. \cite{Oh} and Fox et al. \cite{Fox} have used the same dataset and utilized switching dynamical models to model the six bees dance patterns. Table \ref{C3Tab1} shows accuracy of their approach. Apparently, Parameterized Segmental SLDS proposed by Oh et al. \cite{Oh} works best for modeling the dance patterns for all the bees except \#5. The more accurate method for bee \#5 is Sticky Hidden Markov Model proposed by Fox \cite{Fox}. The mentioned methods are as complicated as they sound and running them requires extensive training time. Besides, they do not support information sharing between the bee dance patterns; each bee requires its own model parameters.

\begin{table}[!hb]
\centering \caption{\label{C3Tab1} Model accuracy}
\begin{tabular}{l c c c c c c} \hline
Bee \# & 1 & 2 & 3 & 4 & 5 & 6 \\ \hline
HDP--VAR(1)--HMM unsupervised & 45.0 & 42.7 & 47.3 & 88.1 & 92.5 & 88.2 \\
HDP--VAR(1)--HMM partially supervised & 55.0 & 86.3 & 81.7 & 89.0 & 92.4 & 89.6 \\
SLDS supervised MCMC & 74.0 & 86.1 & 81.3 & 93.4 & 90.2 & 90.4 \\
PS--SLDS supervised MCMC & 75.9 & 92.4 & 83.1 & 93.4 & 90.4 & 91.0 \\ \hline
\end{tabular}
\end{table}

\subsection{Approach}
Although, the above methods seems to achieve a high model accuracy through sophisticated modeling, we believe a simpler model that runs fast and does not require extensive training is favorable in real applications. This can be achieved through monitoring and then classifying only when a potential change is observed.

\begin{figure}[!ht]
    \centering
    \begin{subfigure}[b]{0.95\textwidth}
        \includegraphics[width=\textwidth]{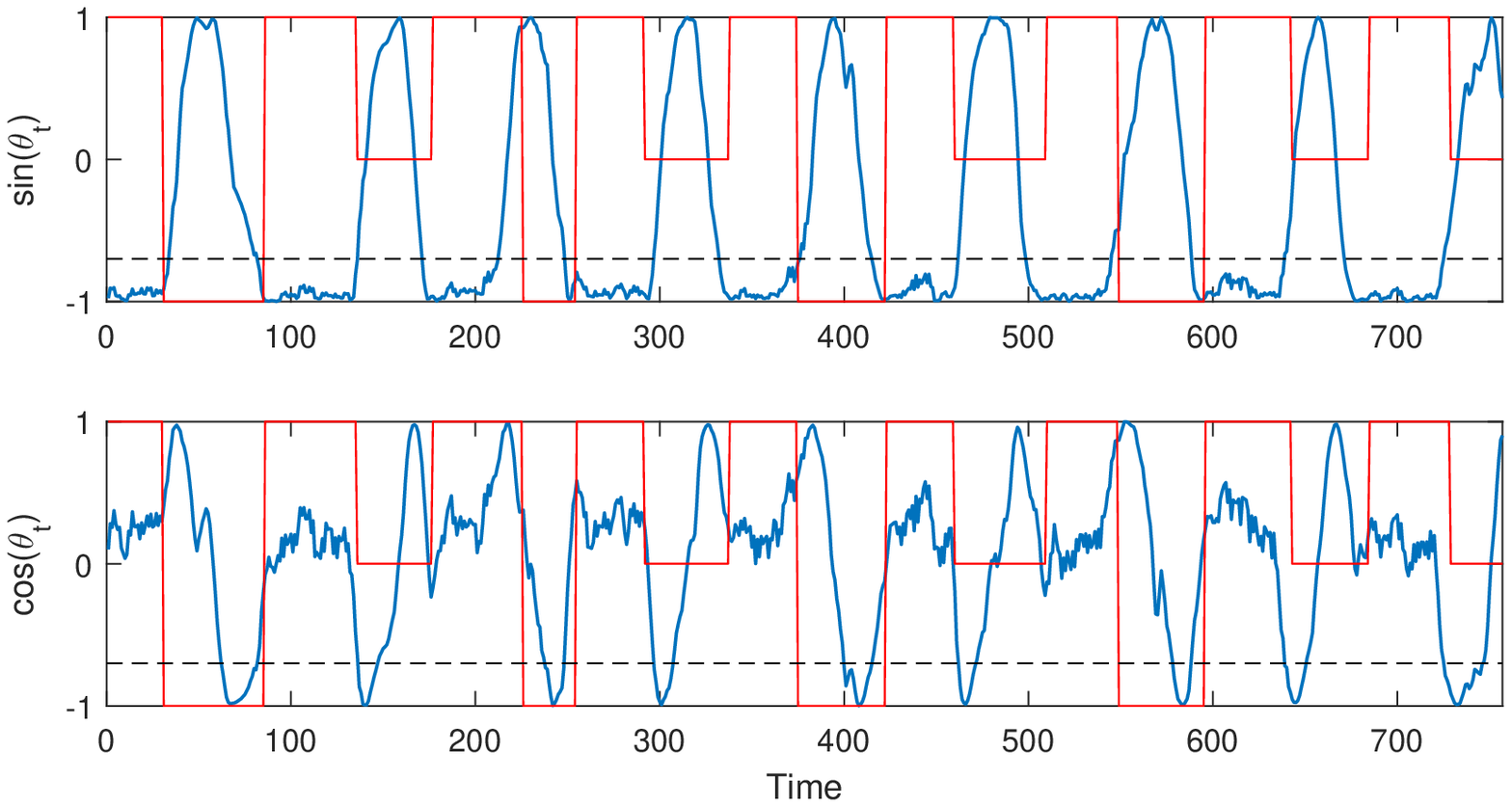}%
        \caption{bee \#4}
    \end{subfigure}
    \begin{subfigure}[b]{0.95\textwidth}
        \includegraphics[width=\textwidth]{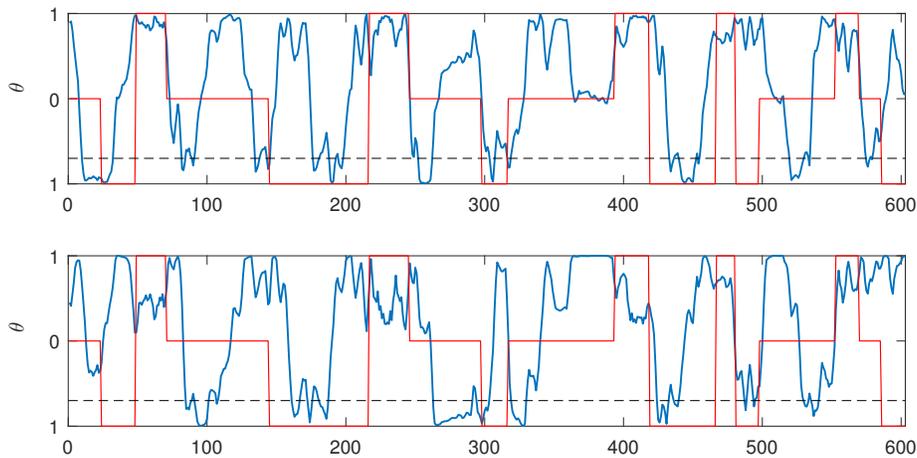}%
        \caption{bee \#3}
    \end{subfigure}
    \caption{Monitoring functions for the bee dance}\label{ch2fig2}
\end{figure}

After investigating some statistical characteristics of the the dance signals such as variance, average, minimum, maximum, moving mean and variance we arrived at the conclusion that a moving average of $\sin(\theta_t)$ with window size 3 can efficiently signal for a change of dance pattern. The moving average of 3 smooths the sharp edges of a signal and helps with detecting steady increase/decrease.
Figure \ref{ch2fig2}.a shows how setting a threshold on $\sin(\theta_t)$ can segment the overall dance signal into intervals of patterns for bee \#4. We set the threshold as -0.7 which could detect all the change dance patterns with some lags at the time they happened. The same pattern could be seen in bee \#5 and \#6 but it was less tractable in bees \#1, \#2, and \#3. The reason could be that each bee possesses her own dancing beat to communicate the same information \cite{Schurch2016} and these three bees danced with more variation than the other three.

By comparing the dance patterns of bee \#3 and bee \#4 in Figure \ref{ch2fig2} we realize that there are some intervals that logically does not match with the provided label for bee \#3. This issue has been mentioned in  \cite{Fox} as well. We could not find out how the labels have been assigned to the overall signal but we could find the videos related to the dance moves. However, the files were not the original recordings and low quality of the videos made it impossible to find the true dance patterns. Thus, we opt to analyze only the dance patterns of bee \#4 to \#6 which seems that have little to no discrepancy.

\begin{figure}[!ht]
    \centering
    \begin{subfigure}[b]{0.44\textwidth}
        \includegraphics[width=\textwidth]{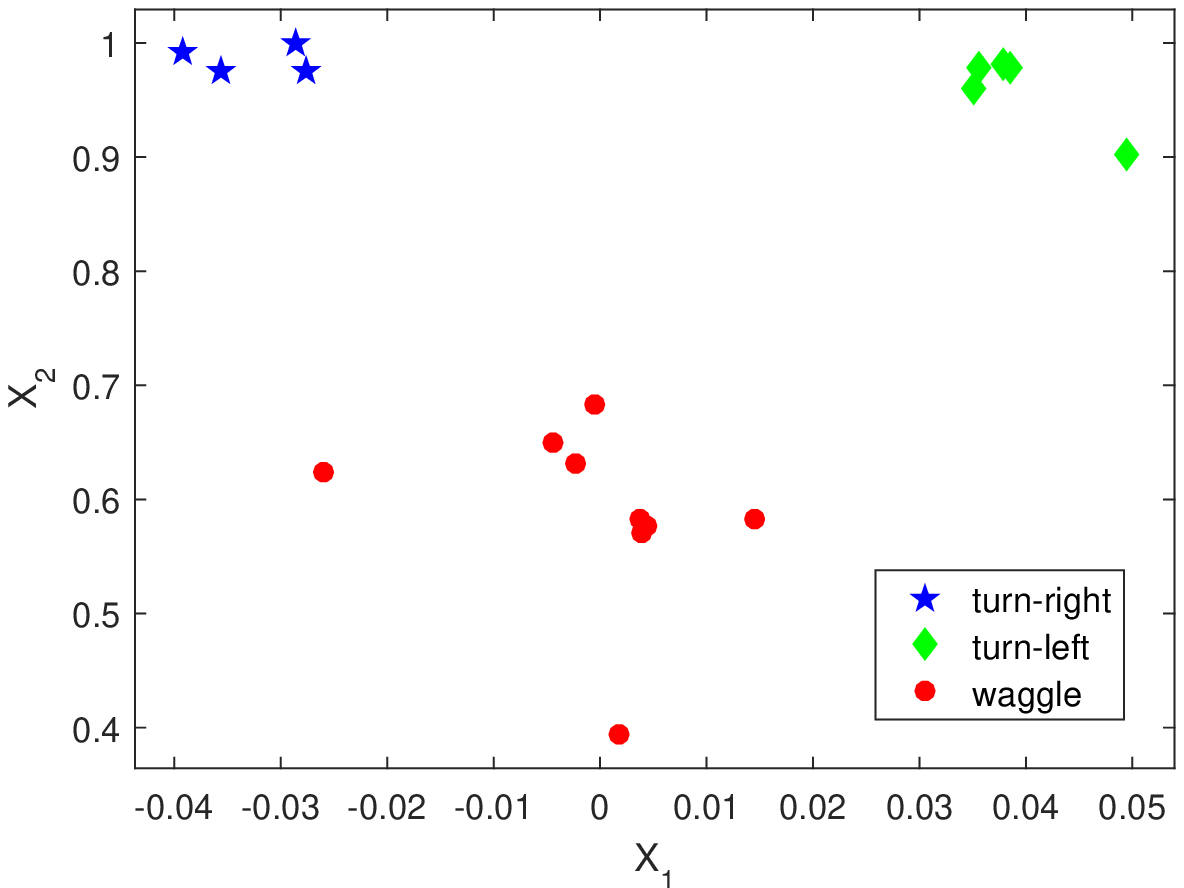}%
        \caption{bee \#4}
    \end{subfigure}
    \begin{subfigure}[b]{0.44\textwidth}
        \includegraphics[width=\textwidth]{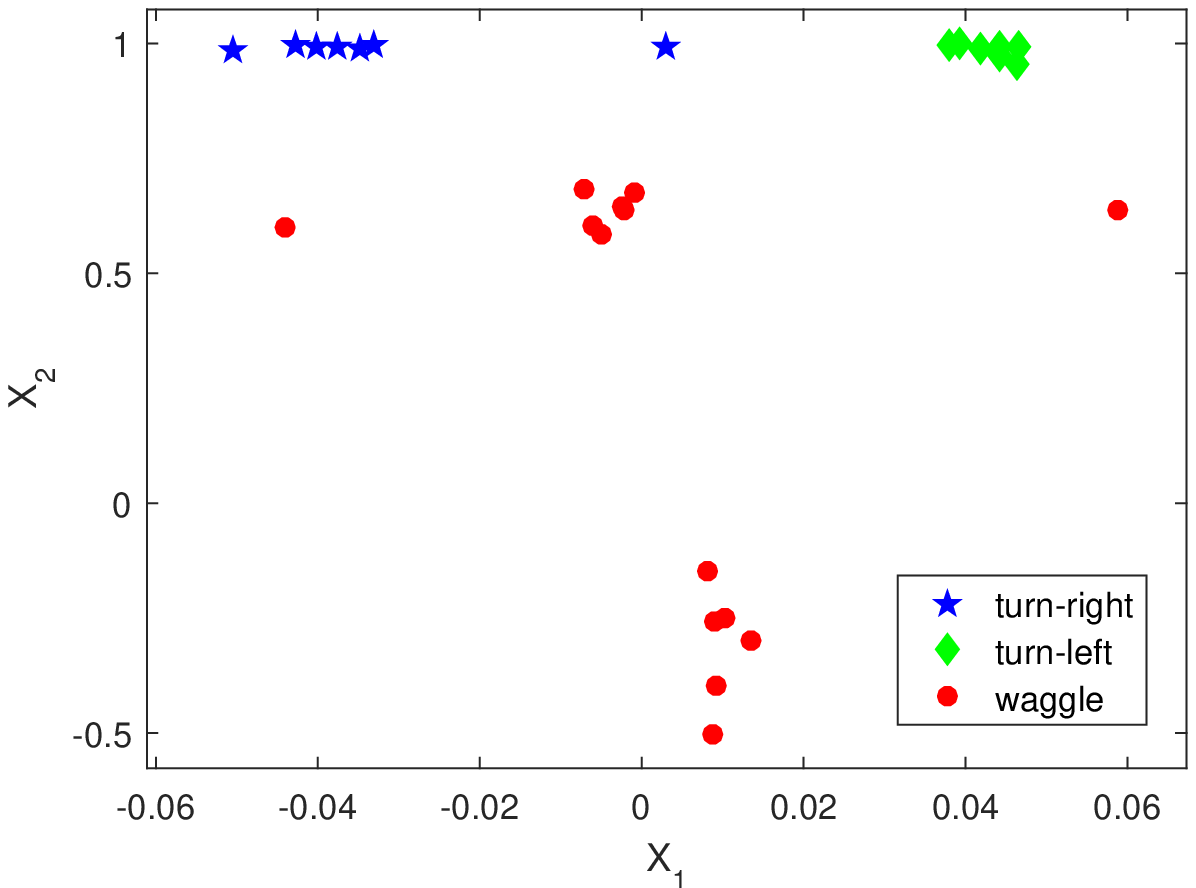}%
        \caption{bee \#5}
    \end{subfigure}
        \begin{subfigure}[b]{0.44\textwidth}
        \includegraphics[width=\textwidth]{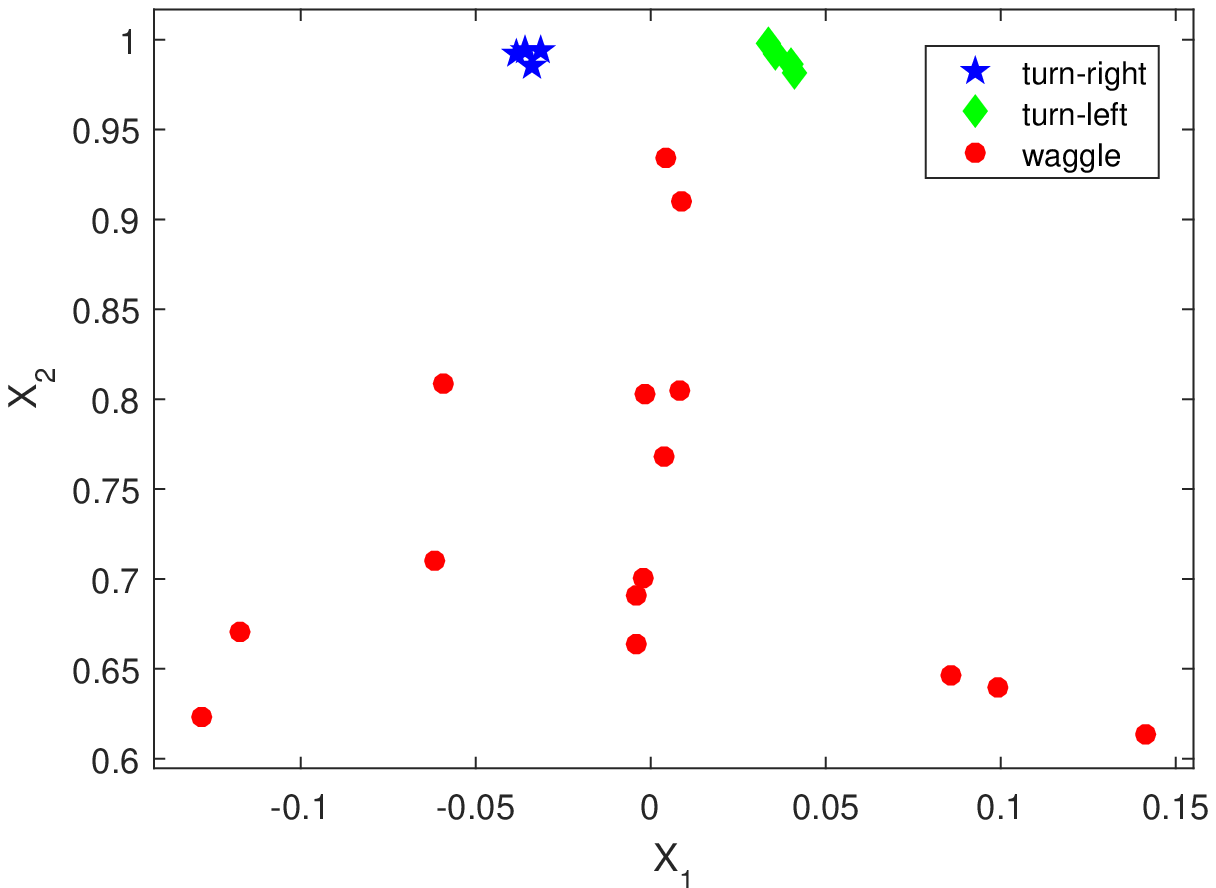}%
        \caption{bee \#6}
    \end{subfigure}
    \caption{The two selected features and their values}\label{ch2fig3}
\end{figure}

Next step is the feature extraction in which we find discriminant features to classify three different dance moves. We started by bee\#4 and after positive feedback applied the same technique to other bees. For bee \#4, after segmenting the continuous signal into 18 intervals of various sizes based on the results of the monitoring section, we extracted various numbers from each interval as potential features. For the head angle $\theta_t$, these features included variance, mean, minimum, and maximum of $\sin(\theta_t)$, $\cos(\theta_t)$, $MovAve(\sin(\theta_t),3)$, $MovAve(\cos(\theta_t),3)$, $Diff(MovAve(\sin(\theta_t),3))$,  $Diff(MovAve(\cos(\theta_t),3))$, etc. computed over a given interval $t\in T$. After investigating the matrix scatter plot of the extracted features, we decided to use $X_1$: mean of the $Diff(MovAve(\cos(\theta_t),3))$ and $X_2$: maximum of $MovAve(\cos(\theta_t),3)$ as features. Figure \ref{ch2fig3}.a best shows the reason for this decision. Obviously, this choice significantly discriminates the three dance patterns.

Going over the same process of monitoring using moving average of $\sin(\theta_t)$, utilizing a threshold of -0.7 for detecting potential change of dance pattern, segmenting the continuous recording signal using the monitoring section, and extracting the two features $X_1$ and $X_2$ for bees \#5 and \#6 will result in scatter plot of features illustrated in Figures \ref{ch2fig3}.b and \ref{ch2fig3}.c. Again, the segmented intervals are well discriminated using the two utilized features. This can result in great classification performance in the next step.

\section{Results}

In this step, the features are fed to three classification algorithms known as Logistic Regression, Artificial Neural Network, and Support Vector Machine. Weka software has been used to perform the classifications with a 5-fold cross validation. Table \ref{C3Tab2} provides the accuracy and F-score of classifying the bee dance patterns using our approach for different bees. The advantages of this approach are

\begin{table}[!hb]
\centering \caption{\label{C3Tab2} Classification Accuracy and F-scoe}
\begin{tabular}{l l c c c c} \hline
 & Bee \# & 4 & 5 & 6 & 4--6 \\ \hline
Logistic Regression & Accuracy &  94.4 & 100. & 87.0 & 97.1 \\
 & F-score &  94.6 & 100. & 87.0 & 97.2 \\ \hline
ANN $2\times3\times3$ & Accuracy &  100. & 100. & 73.9 & 92.9 \\
 & F-score &  100. & 100. & 69.3 & 93.0 \\ \hline
SVM RBF Kernel & Accuracy &  100. & 81.5 & 66.2 & 98.6 \\
 & F-score &  100. & 76.3 & 52.5 & 98.6 \\ \hline
\end{tabular}
\end{table}

\begin{itemize}
  \item High classification accuracy that reaches too 100\% for some cases that beats alternate approaches in the literature.
  \item Fast run time of the algorithm, it runs in fractions of a second, which makes it suitable for real-time and real-life predictions.
  \item Simplicity of the proposed process
  \item Possibility of pooling the samples together for higher classification accuracy.
\end{itemize}
As seen in the table, we can pool the extracted features together and then perform classification. This improves the accuracy of the results since there are more training samples.

\section{Conclusion and Discussion}

\begin{figure}[!ht]
\centering
\includegraphics[width=1\textwidth,height=1\textheight,keepaspectratio]{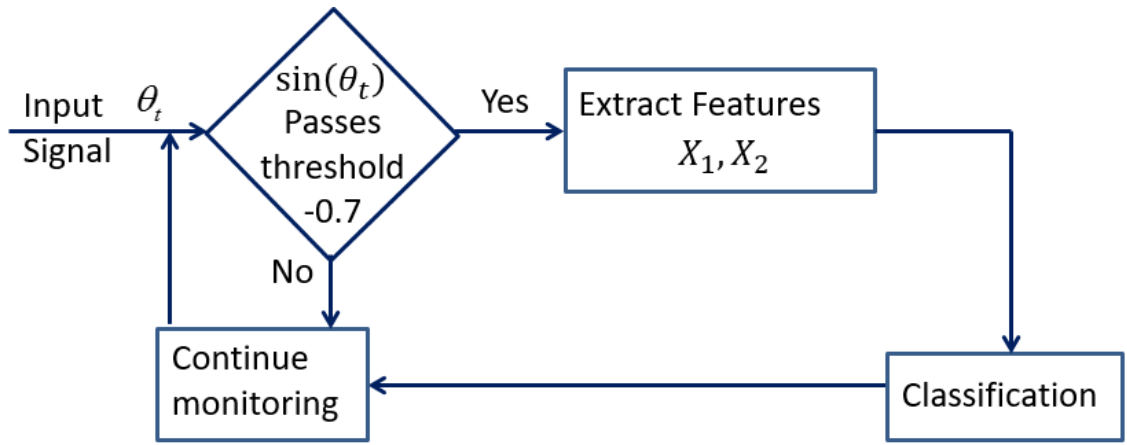}%
\caption{Real-time application circuit} \label{ch2fig4}
\end{figure}

The overall decision circuit is given in Figure \ref{ch2fig4}. First, the head angle $\theta_t$ signal is monitored in real time by $\sin(\theta_t)$ computed at 30 Hz frequency. Upon passing a threshold of -0.7, the last three seconds of signal is used to extract two features $X_1, X_2$. The features are then fed to a selected classifier to predict the type of dance move. The monitoring then continues. The procedure needs calibration before applying.

As seen in this application, the monitoring-action process significantly improves the processing time of classifying signals. The processing time is a huge issue with machine learning analytics. We are currently working to present the results of this chapter to a peer reviewed journal.


\end{document}